\documentclass[11pt,a4paper,dvipsnames]{article}
\usepackage[hyperref]{acl2021}
\usepackage{times}
\usepackage{latexsym}

\usepackage{microtype}

\aclfinalcopy 
\def\aclpaperid{3238} 

\usepackage[linesnumbered, ruled]{algorithm2e}
\usepackage[T1]{fontenc}
\usepackage{dsfont}
\usepackage{xcolor}

\usepackage{mdframed}
\usepackage{booktabs}
\usepackage{graphicx}

\usepackage[normalem]{ulem}
\usepackage{enumitem}
\usepackage{soul}
\colorlet{soulred}{red!30}
\sethlcolor{soulred}%

\usepackage{amsmath,amsthm,amsfonts,amssymb,bm}
\usepackage{framed}
\usepackage{varioref}
\usepackage{float}
\usepackage{multirow}
\usepackage{dashrule}
\usepackage{url}

\usepackage[T5,T1]{fontenc}
\usepackage[utf8]{inputenc}
\DeclareTextSymbolDefault{\ohorn}{T5}
\DeclareTextSymbolDefault{\uhorn}{T5}

\usepackage{pifont}

\newcommand*{\SuperScriptSameStyle}[1]{%
  \ensuremath{%
    \mathchoice
      {{}^{\displaystyle #1}}%
      {{}^{\textstyle #1}}%
      {{}^{\scriptstyle #1}}%
      {{}^{\scriptscriptstyle #1}}%
  }%
}

\newcommand*{\sig}{\SuperScriptSameStyle{*}}

\title{Language Embeddings for Typology and Cross-lingual Transfer Learning}

\author{Dian Yu\thanks{\; Equal contribution.}, \hspace{1ex}Taiqi He\footnote[1]{} \hspace{1ex}and Kenji Sagae\\
~~\\
  University of California, Davis \\
  \texttt{\{dianyu, tqhe, sagae\}@ucdavis.edu} \\}

\date{}

\begin{document}
\maketitle
\begin{abstract}
Cross-lingual language tasks typically require a substantial amount of annotated data or parallel translation data. We explore whether language representations that capture relationships among languages can be learned and subsequently leveraged in cross-lingual tasks without the use of parallel data. We generate dense embeddings for 29 languages  using a denoising autoencoder, and evaluate the embeddings using the World Atlas of Language Structures (WALS) and two extrinsic tasks in a zero-shot setting: cross-lingual dependency parsing and cross-lingual natural language inference\footnote{Our learned language embeddings and code available at \url{https://github.com/DianDYu/language_embeddings}}. 
\end{abstract}

\section{Introduction}

Recent efforts to leverage multilingual datasets in language modeling \cite{xlm, devlin-etal-2019-bert} and machine translation \cite{johnson-etal-2017-googles, lu-etal-2018-neural} highlight the potential of multilingual models that can perform well across various languages, including ones for which training sets are scarce.
Most of the current multilingual research focuses on learning invariant representations or removing language-specific features after training~\citep{libovicky-etal-2020-language, bjerva-augenstein-2021-typological}. 
Despite recent advances, there are still limitations. Previous work has shown that similar languages can benefit from sharing parameters, but less similar languages do not help \cite{transfer_MT, pires-etal-2019-multilingual}. However, in spite of some interests in typology~\citep{ponti-etal-2019-modeling}, identifying similar languages is nontrivial, especially for less studied ones. In addition, as \citet{theoretical_zhao} suggest, learning invariant representations can actually harm model performance.
Therefore, in order to leverage language agnostic and language specific information effectively, we propose to generate language representations and examine the interactions among different language representations.

One way to represent language identity within a multilingual model is the use of language codes, or dense vectors representing language embeddings. If languages are represented with vectors that capture cross-lingual similarities and differences across different dimensions, this information can guide a multilingual model regarding what and how much of the information in the model should be shared among specific languages. 
Much of the previous research focused on generating language embeddings using prior knowledge such as word order \cite{ammar-etal-2016-many, littell_lang_emb}, using a parallel corpus~\citep{bjerva-etal-2019-language, ostling-tiedemann-2017-continuous}, and using
language codes as an indicator to distinguish input and output words in a shared vocabulary into different languages \cite{johnson-etal-2017-googles, xlm}. In contrast, our work focuses on generating and using language embeddings more 
effectively 
as soft-sharing \cite{dependency_parsing} of parameters among various languages in a single model. Furthermore, we are motivated by a more difficult setting where the properties of each language are not known in advance, and no parallel data is available.

We investigate whether we can generate language embeddings to represent typological information
derived solely from corpora in each language without the use of any parallel text, translation models, or prior knowledge. Inspired by the findings that structural similarity, especially word ordering, is crucial in large pretrained multilingual language models \cite{k2019crosslingual}, we propose an unsupervised method leveraging denoising autoencoders \cite{vincent2008extracting} to generate language embeddings. 
We show that our approach captures typological information
by comparing the information in our language embeddings to language-specific information in the World Atlas of Language Structures (WALS, \citealp{wals}).
In addition, to address the question of whether the learned language embeddings can help in downstream language tasks, we plug-in the language embeddings to cross-lingual dependency parsing and natural language inference (XNLI, \citealp{conneau-etal-2018-xnli}) in a zero-shot learning setting, obtaining performance improvements.

\section{Related Work}

Previous related research approached language representations by using prior knowledge, dense language embeddings with multilingual parallel data, or no prior knowledge about languages but having language embeddings primarily as a signal to identify different languages.

\subsection{Feature-based language representations}
An intuitive method to represent language information is through explicit information such as known word order patterns \cite{ammar-etal-2016-many, little-2017-connecting}, part-of-speech tag sequences \cite{wang_typology_prediction}, and syntactic dependencies \cite{ostling-2015-word}. 
\citet{littell_lang_emb} propose sparse vectors using pre-defined language features such as known typological and geographical information for a given language. However, linguistic features may not be available for less studied languages.
Our proposed approach assumes no prior knowledge about each language, deriving typological information from plain text alone. Once a vector for a target language is created, it contains many typological features of the target language, and can be used for transfer learning in downstream tasks.

\subsection{Dense representation with parallel data}
Other previous work has also explored dense continuous representations of languages. One method is to append a language token to the beginning of a source sentence and train the language embeddings with a many-to-one neural machine translation model~\citep{malaviya-etal-2017-learning, tan-etal-2019-multilingual}. Another method is to concatenate language embedding vectors to a character level language model~\citep{ostling-tiedemann-2017-continuous, bjerva-augenstein-2018-phonology, bjerva-etal-2019-probabilistic}. These two methods require parallel translation data such as Bible and TED Talk. \citet{rabinovich-etal-2017-found} derive typological information in the form of phylogenetic trees from translation of different languages into English and French using the European Parliament speech corpus \cite{europarl}, based on the assumption that unique language properties are present in translations \cite{remain1, remain2}. \citet{bjerva-etal-2019-language} abstract the translated sentences from other languages to English with part-of-speech tags, function words, dependency relation tags, and constituent tags, and train the embedding vectors by concatenating a language representation with a symbol representation. In comparison, we generate our language embeddings using no parallel corpora or linguistic annotation, which is suitable for a wider variety of languages, including in situations where no parallel data or prior knowledge is available.

\subsection{Language vectors without parallel data}
The approach that is closest to ours is XLM~\citep{xlm}, which adds language embeddings to each byte pair embedding using Wikipedia data in various languages with a masked language modeling objective. However, similar to \citet{johnson-etal-2017-googles}, the trained language embeddings only serve as an indicator to the encoder and decoder to identify input and output words in the vocabulary as belonging to different languages. In fact, in a follow up paper, XLM-R \cite{conneau-etal-2020-unsupervised}, language embeddings are removed from the model for better code-switching, which suggests that the learned language embeddings may not be optimal for cross-lingual tasks. In this paper, following the finding that structural similarity is critical in multilingual language models \cite{k2019crosslingual}, we generate language embeddings from a denoising autoencoder objective and demonstrate that they can be effectively used in cross-lingual zero-shot learning.

\section{Generating Language Embeddings}

We first present the data used to generate language embeddings, then introduce our approach inspired by denoising autoencoders \cite{vincent2008extracting}.

\subsection{Data and preprocessing} \label{preprocessing}

To train our multilingual model, we use the CommonCrawl dataset from the CoNLL 2017 shared task \cite{conll_2017_auto_data}
to obtain monolingual plain text in various languages.
To represent words across different languages in a shared space, we use the unsupervised pretrained
aligned word embeddings from MUSE \cite{facebook_muse}. We choose the 29 languages from the CoNLL 2017 monolingual text dataset for which
MUSE pretrained embeddings are available.\footnote{\url{https://github.com/facebookresearch/MUSE}} 
A subset of 200K sentences are selected randomly for each language. The languages we use are: English, French, Romanian, 
Arabic, German, Russian, 
Bulgarian, Greek, Slovak,  
Catalan, Hebrew, Slovene,
Croatian, Hungarian, Spanish,
Czech, Indonesian, Swedish, 
Danish, Italian, Turkish,
Dutch, Norwegian Bokmål, Ukrainian,
Estonian, Polish, Vietnamese,
Finnish, and Portuguese, which cover ten language genera.

We experiment with two types of word representations in training language embeddings. The most straightforward way is to use the pretrained MUSE embedding for each specific language (we refer to this setting as \textbf{Spe.}).
We also experimented with mapping word embeddings from different languages into one language (English in our experiments because it is used as the pivot language in MUSE embeddings, \textbf{Eng.}) for three reasons. First, because MUSE is mainly trained by an orthogonal rotation matrix and the distances among words in each language are still maintained thereafter, language identities can potentially be revealed. The result is that the learned language embeddings reflect the features incorporated in the unsupervised word mapping methods instead of the intrinsic language features. Second, 
we hypothesize that mapping to a single language space requires the model to encode more information in language embeddings as their language identities instead of relying on their revealed ones. Finally, using shared word embeddings can reduce the vocabulary size for memory concerns by effectively reducing both the lookup table size and the output softmax dimension size. 

For \textbf{Eng.} word embedding mapping, we align words from different languages to English embeddings using cross-domain similarity local scaling (CSLS, \citealp{facebook_muse}).
The vocabulary of our model is restricted to the words in the English MUSE embeddings, and all unknown words are replaced with a special unknown token.
Although imperfect mapping from each language to English tokens may introduce noise (see scores in Appendix \ref{muse_word}) and result in a coarse approximation of the original sentences, crucial syntactic and semantic information should still be present. 

In our experiments, a language code is appended to each token according to the original language of the sentence.
For instance, the German sentence ``Er hat den roten Hund nicht gesehen" would be represented in our \textbf{Spe.} condition as 
\begin{equation}
    \nonumber
    \small
    \text{\color{red}{Er}\_\color{OliveGreen}{de} \color{red}{hat}\_\color{OliveGreen}{de} \color{red}{den}\_\color{OliveGreen}{de} \color{red}{roten}\_\color{OliveGreen}{de} \color{red}{Hund}\_\color{OliveGreen}{de} \color{red}{nicht}\_\color{OliveGreen}{de} \color{red}{gesehen}\_\color{OliveGreen}{de}}
\end{equation}
and in the \textbf{Eng.} condition as 
\begin{equation}
    \nonumber
    \small
    \text{\color{blue}{he}\_\color{OliveGreen}{de} \color{blue}{has}\_\color{OliveGreen}{de} \color{blue}{the}\_\color{OliveGreen}{de} \color{blue}{red}\_\color{OliveGreen}{de} \color{blue}{dog}\_\color{OliveGreen}{de} \color{blue}{not}\_\color{OliveGreen}{de} \color{blue}{seen}\_\color{OliveGreen}{de}}
\end{equation}
Intuitively, the idea is to have the words themselves be the same across languages (either through the aligned MUSE embeddings or by direct mapping to English words), and let the additional language code provide to the model the information that would explain the structural differences observed across languages in the training data.

\subsection{Denoising autoencoder}
Given a multilingual plain text corpus with sentences in each language (and no parallel text), we first perturb each sentence to create a noisy version of the sentence where its words are randomly shuffled. The training objective is to recover the original sentences, which requires the model to learn how to order words in each language. We hypothesize that compared to language modeling, this will encourage the language embeddings to learn more structural information instead of relying on topics or word co-occurrence to generate meaningful training sentences.
We implement our multilingual denoising autoencoder with an LSTM-based \cite{lstm} sequence-to-sequence model \cite{seq2seq}. The input strings are perturbed sentences and the output strings are the original sentences. 
See Appendix \ref{training_details} for implementation details.

After preprocessing the data,
we concatenate a language embedding vector initialized from normal distribution as a language identity feature (the language code mentioned in Section \ref{preprocessing}) to each of the pretrained word embeddings.
Since certain languages are more similar to, or more different from, each other,  the model will learn how to reorder a sequence of words depending on the specific language.
For example, reordering an Italian sentence should be more similar to reordering a Spanish sentence than it is to reordering a German sentence. 
Because the decoder captures the actual word order of the sentences in each target language, whereas the language codes in the encoder are meant to capture only language identity and no word order information, we use the extracted language embeddings from the decoder in our experiments.\footnote{To confirm our assumption about the embeddings for the language codes in the encoder and the decoder, we also performed experiments using the encoder language embeddings. As expected, the results obtained with embeddings from the encoder were inferior in every case tested.}
Each word is represented with a pretrained 300-dimensional vector, and each language embedding is represented with a 50-dimensional vector\footnote{We experimented with different dimensions for language embedding and did not observe performance difference.}. 
The input token is thus a 350-dimensional vector from the concatenation.


\section{Experiments}
To examine the quality of the typological information captured by the language embeddings, we perform intrinsic and extrinsic evaluations. Our intrinsic evaluation consists of predicting linguistic typology and language features from the World Atlas of Language Structures (WALS, \citealp{wals}).
Our extrinsic evaluations are based on cross-lingual dependency parsing and cross-lingual natural language inference (XNLI, \citealp{conneau-etal-2018-xnli}) in a zero-shot learning setting, where a trained model makes predictions on a language not seen during training, but for which a language embedding has been learned from plain monolingual text. 
In contrast with previous research which applies learned typology to cluster similar languages and train machine translation tasks in clusters~\citep{tan-etal-2019-multilingual}, we explore if we can apply the learned embeddings directly into downstream tasks.
We compare three different sets of embeddings based on our approach with three sets of embeddings from previous work:\\
\indent \textbf{Spe. lang\_emb} represents language embeddings from our proposed denoising autoencoder trained with language specific MUSE embeddings, using CommonCrawl text.
\\
\indent \textbf{Eng. lang\_emb} represents language embeddings trained with English MUSE embeddings after mapping words from different languages to English, using CommonCrawl text.\\
\indent \textbf{Wiki lang\_emb} represents language embeddings trained with English MUSE embeddings using Wikipedia. We use the same data selection and preprocessing process as detailed in Section \ref{preprocessing}. We use these embeddings to show the impact of training data. In addition, we use these embeddings to compare with XLM embeddings trained with Wikipedia.\\
\indent \textbf{Malaviya} represents language embeddings from \citet{malaviya-etal-2017-learning}, trained with a many-to-one machine translation model using Bible parallel data. It has 26 languages in common with our 29 languages except English, Hebrew, and Norwegian. We use these embeddings to represent previous methods of learning language representations from parallel data.\footnote{We do not evaluate the embeddings from \citet{malaviya-etal-2017-learning} on parsing and XNLI because they do not include English embeddings, which are necessary for a direct comparison. In XNLI, in particular, there is only training data for English.}\\
\indent \textbf{XLM mono} represents language embeddings trained with XLM model using the same monolingual data as Wiki lang\_emb on 29 languages.\\
\indent \textbf{XLM parallel} represents language embeddings trained with XLM using monolingual and parallel data from 15 XNLI languages. We extract the embeddings from the publicly available model.\\


\subsection{Linguistic typology prediction}
We first inspect the language embeddings qualitatively through principle component analysis (PCA) visualization. We also use spectral clustering to recover the language genus (language family subgroup) information from the embeddings. To compare the quality of the clusterings quantitatively, we calculate the adjusted Rand index \cite{hubert1985rand} between the generated clusters and the actual language genera. 

\subsection{WALS feature prediction}
We evaluate the language embeddings on predicting language features in WALS. Each WALS feature describes a characteristic of languages, such as the order of subject, object, and verb. We consider the features for which information is available for more than 50\% of the languages we use and cast each feature prediction as a multi-class classification task. We then classify the features into the following categories (see details in Appendix \ref{app_wals_details}).
\begin{itemize}
	\item \textbf{Lexicon}: usage of specific words, e.g. whether the language has  separate words for ``hand'' and ``arm'', etc.;
	\item \textbf{Syntax}: mostly related to the relative orders between various types of constituents, including order of subject, object and verb, adpositions and noun phrases, and also features related to syntactic constructions;
	\item \textbf{Partially Morphological (Part. Morph.)}: features that mainly concern syntax or semantics but either usually relate to morphology (such as inflectional morphemes), or have morphological information coded in the values of the features, e.g. gender systems, order of negative morphemes and verbs;
	\item \textbf{Non-learnable}: features that mainly concern morphology, phonology, or phonotactics, and are not learnable from reordering plain text.
\end{itemize}

The categories make it easier to evaluate what the language embeddings capture.
We train linear classifiers to predict WALS results. For each feature, we hold out one language and train a classifier on the language embeddings of the rest of the languages to predict the corresponding feature values on the held-out language embedding, in a leave-one-out cross-validation scheme. We then average the accuracy of the features within each category to report the results. In addition to comparing different language embeddings, we also compare to two baselines: a \textbf{Random} baseline, and a \textbf{Majority} baseline (which predicts the most common value for each feature). We repeat this procedure $100$ times while randomly permuting the orders of the input vectors to the classifiers to eliminate possible effects due to initial states and report the average and significant scores.

Compared to a recent shared task where the input is some features of a language (e.g. language family and various WALS features), with optionally pre-computed language embeddings to develop models to predict other features \cite{bjerva-etal-2020-sigtyp}, we investigate if trained language embeddings alone can be used to predict WALS features. In addition, we showed that our language embeddings outperformed a frequency baseline among other baselines (see Section \ref{sec:wals_results}) compared to \citet{bjerva-etal-2020-sigtyp}.

\begin{figure*}[h]
\centering
\includegraphics[scale=0.53]{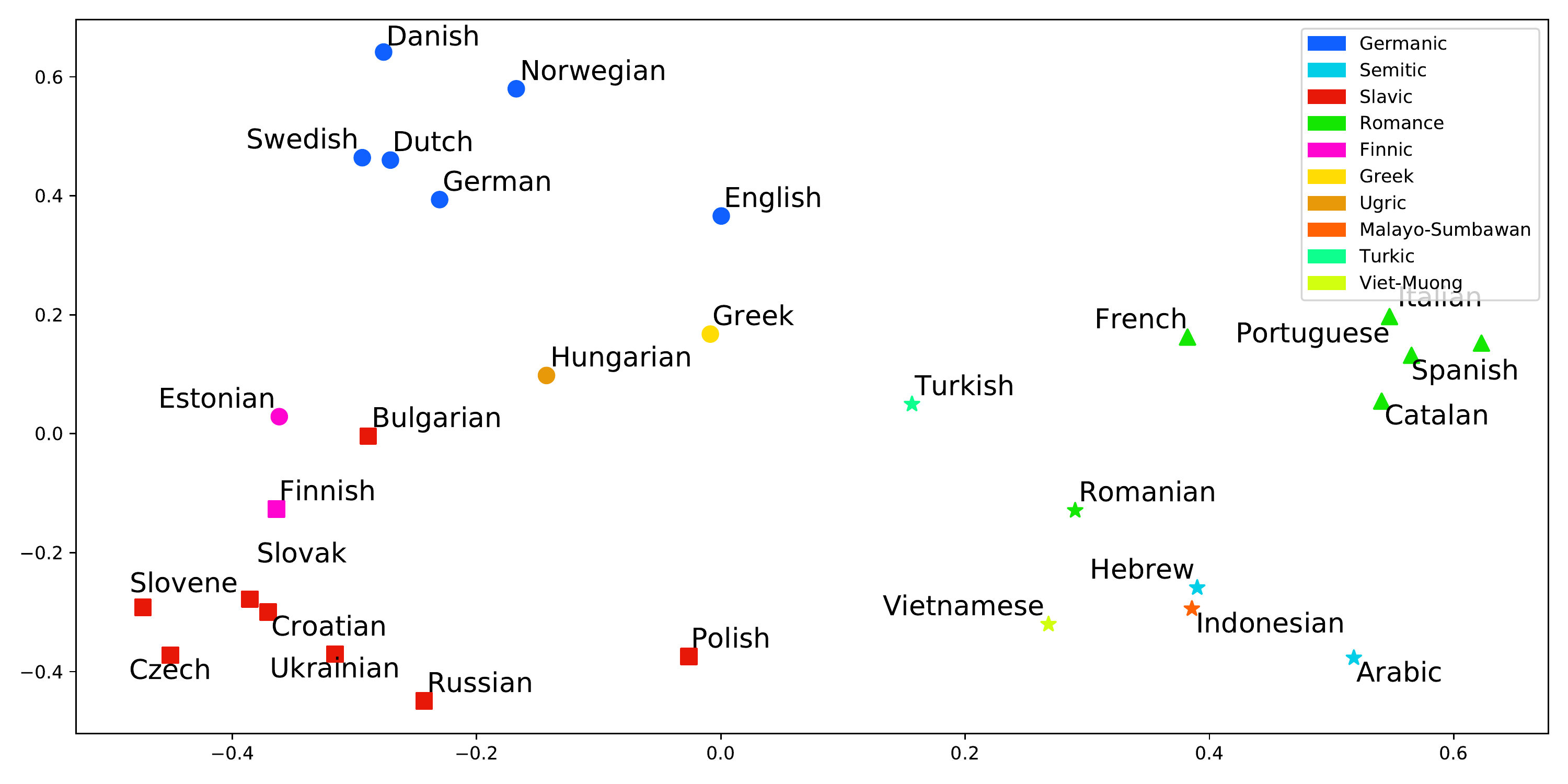}
\caption{\label{pca} Two-dimensional PCA projection of the 50-dimensional language embeddings. Shapes represent automatically derived clusters, and colors represent language genera.}
\end{figure*}

\subsection{Cross-lingual dependency parsing}

Since our language embeddings are trained using a word ordering task, we hypothesize
that they capture syntactic information. To verify that meaningful syntactic information
is captured in the language embeddings, we use a dependency parsing task where
sentences for each target language are parsed with a model trained with treebanks
from other languages, but no training data for the target language. This can be seen
as a form of cross-lingual parsing or zero-shot parsing, where multiple source languages are used to train a model for a new target language.
Without annotated training data for parsing a target language, the model is expected to leverage treebanks from other languages through language embeddings.

We use 16 languages from Universal Dependencies v2.6~\citep{universal_dependencies}, representing five distinct language genera
(Table~\ref{tab:dependency_results}). 
We modified Yu Zhang's implementation\footnote{\url{https://github.com/yzhangcs/parser}} of biaffine dependency parser~\cite{dozat_deep_2016}.
In specific, we freeze word embeddings, concatenate a 50-dimensional embedding (either the corresponding
Eng. language embedding or a random embedding) to the embedding of each token, and not use part-of-speech
information (since we are assuming no annotated data is available for the target
language).
The goal of this evaluation
is not to obtain state-of-the-art attachment scores, but to find whether a model
that uses our language embeddings produces higher attachment scores than a model that
instead uses random embeddings of the same size\footnote{Random embeddings are used to eliminate the effect of different dimensionality. In our preliminary experiments, we found that adding a random embedding performs better than not adding any embedding.}. While our embeddings should
capture syntactic typology, random embeddings would simply indicate to the model the language for each sentence
with no information about how languages are related.

\subsection{XNLI}
Natural language inference (NLI) is a language understanding task where the goal is to predict textual entailment between a premise and a hypothesis as a three-way classification: \emph{neutral}, \emph{contradiction}, and \emph{entailment}. 
The XNLI dataset \citep{conneau-etal-2018-xnli} translates English NLI validation and test data into 14 other languages. We evaluate on ten of the XNLI languages which we trained language embeddings with.

State-of-the-art models on XNLI are Transformers~\citep{transformer} pretrained on large corpora~\citep{xtreme}. To evaluate if our learned language embeddings (from an LSTM model) can be plugged  off-the-shelf into other architectures such as Transformer, we compare with two strong Transformer-based baselines, XLM (\citealp{xlm}. L = 12, H = 1024, 250M params) and XLM-R (\citealp{conneau-etal-2020-unsupervised}. XLM-R\textsubscript{Base}: L = 12, H = 768, 270M params; XLM-R\textsubscript{Large}: L = 24, H = 1024, 550M params). XLM 
adds language embeddings together with each word embedding and position embedding as the input embedding in training masked language modeling (MLM, with monolingual data) and/or a translation language modeling (TLM, with translation parallel data). In comparison, XLM-R removes language embedding and is pretrained with MLM on much more data. We train our model on the English MultiNLI~\citep{williams-etal-2018-broad} dataset, and directly evaluate the trained model on the other languages without language-specific fine-tuning, in a zero-shot cross-lingual setting. To select the best checkpoint for test set evaluation, we follow \citet{conneau-etal-2020-unsupervised} by evaluating on the development set of all languages. In addition, we also experiment with a fully zero-shot transfer setting where we select the best checkpoint by evaluating on the English development set. We run the selected checkpoint on the test set of each language and report the accuracy scores. We use the public available XLM model pretrained on 15 XNLI languages with MLM and TLM objectives, and XLM-R pretrained on 100 languages.
In order to add our learned language embeddings into XLM and XLM-R models, we normalize our embeddings to have the same variance as the XLM language embeddings, and we learn a simple linear projection layer to map our 50-dimension embeddings (which is frozen during training) to the hidden dimension of corresponding models. We report all results averaged over three random seeds. 
See Appendix \ref{xnli_details} for implementation details.


\section{Results and Analysis}

We show results of our proposed language embeddings in comparison to the baselines and language vectors generated from previous work on linguistic typology, WALS, cross-lingual parsing, and XNLI. We report results with \textbf{Eng.} language embeddings. Detailed comparison to other language embeddings on each task can be found in Appendix \ref{app_addtional_results}.



\subsection{Linguistic Typology}
\label{sec:typology}
\begin{table}[h]
\begin{center}
\resizebox{\columnwidth}{!}{
\begin{tabular}{l|cccc|c}
\toprule

& Lexicon & Syntax & Part. Morph. & Non-learn. & Rand \\

$n$ features & 2 & 14 & 46 & 20 & - \\
\midrule
Random  & 0.56 & 0.61 & 0.52 & 0.52 & -  \\
Majority  & 0.64 & 0.75 & 0.69 & 0.68 & - \\
\midrule
Malaviya & 0.66\sig & 0.74 & 0.66 & 0.66 & 0.13\\
XLM mono & 0.41 & 0.75 & 0.66 & 0.68 & 0.12\\
\midrule
Spe. & 0.64 & 0.78\sig & 0.68 & 0.66 & 0.53\\
Eng. & 0.85\sig & 0.79\sig & \textbf{0.71\sig} & 0.66 & \textbf{0.58}\\
Wiki & \textbf{0.87\sig} & \textbf{0.81\sig} & 0.70\sig & \textbf{0.68} & 0.51\\

\bottomrule
\end{tabular}
}
\end{center}

\caption{\label{wals_results} WALS prediction and linguistic typology clustering results on 26 in-common languages across 10 language genera. 
\sig indicates statistical significance ($p<0.01$) over the Majority baseline.}
\end{table}

Figure \ref{pca} shows a two-dimensional PCA projection of the learned language embeddings. Due to space limitations, we only show the projection of the language embeddings using words mapped to English embeddings; using language-specific embeddings produces similar results. We can clearly see the clustering of Slavic languages on the lower left, Romance on the right, and Germanic on the upper left. Our dataset also contains two Finnic languages, which appear right above the Slavic languages, and two Semitic languages, which appear on the lower right. The other languages, Vietnamese, Indonesian, Turkish, and Greek,
are from language groups underrepresented in our dataset, and appear either mixed with the Germanic languages (in the case of Hungarian, Turkish and Greek), or far on the lower right corner (Vietnamese, Indonesian). Romanian, a Romance language, appears miscategorized by our language embeddings. While it is close to the cluster of romance languages, it appears closer to the singleton languages in the dataset and to the two Semitic languages.

In addition to actual language relationships represented by color, we also present the result of spectral clustering with four categories through different shapes.
Results illustrate that our language embeddings can capture similarities and dissimilarities among language families. In comparison, language embeddings generated by \citet{malaviya-etal-2017-learning} do not capture clearly visible language relationships (see Appendix \ref{app_topology}). Quantitatively, clusters from our learned language embeddings (\textbf{Eng.}) achieve a much higher Rand score (0.58) compared to previous language embeddings, as shown in Table \ref{wals_results} (last column). This indicates that our clusters closely align with true language families. 


\subsection{WALS predictions}
\label{sec:wals_results}


Table \ref{wals_results} shows the prediction accuracy for WALS features, averaged within each category. Unlike the language representations generated by \citet{bjerva-etal-2019-language}, which do not outperform the majority baseline without finetuning, our derived language embeddings perform significantly better than the baselines and previous methods in lexicon, syntax, and partially morphological categories. 
Note that even though the training objective of the denoising autoencoder is to recover a language-specific word order, the model does not use linguistic features such as grammatical relation labels or subject-verb-object order information. Instead, it derives typological information from text alone through the word reordering task. The language embeddings generated with words mapped to English embeddings (\textbf{Wiki} and \textbf{Eng.}) generally produce more accurate predictions, with the models trained from Wikipedia producing slightly better results likely due to cleaner training data. We show WALS results comparison on 29 languages and comparison to XLM parallel in Appendix \ref{app_wals}. Results from different settings show that we do not need clean data (e.g. Wiki) to generate language embeddings.

\begin{table}[h]
\resizebox{\columnwidth}{!}{
\begin{tabular}{lcc}
\toprule
Language & Baseline & Language Emb. \\
\midrule
\multicolumn{3}{l}{\textit{Finnic}} \\
\quad Estonian & 56.19 & \textbf{61.68} \small{(+5.49)}\\
\quad Finnish & 59.59 & \textbf{62.91} \small{(+3.32)}\\
\multicolumn{3}{l}{\textit{Germanic}} \\
\quad Danish & 63.31 & \textbf{69.62} \small{(+6.31)}\\
\quad English & \textbf{74.51} & 74.08 \small{(-0.43)}\\
\quad German & 64.36 & \textbf{65.67} \small{(+1.31)}\\
\quad Norwegian & 77.19 & \textbf{78.20} \small{(+1.01)}\\
\quad Slovene & 67.92 & 67.91 \small{(-0.01)}\\
\multicolumn{3}{l}{\textit{Romance}} \\
\quad Catalan & 72.41 & \textbf{80.76} \small{(+8.35)}\\
\quad French & 68.75 & \textbf{79.37} \small{(+10.62)}\\
\quad Spanish & 74.42 & \textbf{81.74} \small{(+7.32)}\\
\quad Portuguese & 71.11 & \textbf{79.57} \small{(+8.46)}\\
\multicolumn{3}{l}{\textit{Semitic}} \\
\quad Arabic & 48.44 & \textbf{52.51} \small{(+4.07)} \\
\quad Hebrew & \textbf{41.87} & 33.66 \small{(-8.21)}\\
\multicolumn{3}{l}{\textit{Slavic}} \\
\quad Bulgarian & 62.91 &\textbf{ 67.00} \small{(+1.09)}\\
\quad Czech & 65.62 & \textbf{66.98} \small{(+1.36)}\\
\quad Russian & 62.10 & \textbf{66.45} \small{(+4.35)}\\
\midrule
Average & 64.61 & \textbf{68.01} \small{(+3.40)} \\
\bottomrule
\end{tabular}
}
\caption{\label{tab:dependency_results} Zero-shot parsing results (UAS), where
each of 16 languages are parsed using annotated language from the other
15 languages. In the \textit{Language Emb.} column, results were obtained by
concatenating the language embedding to each token's MUSE embedding. In the 
\textit{Baseline} column, results were obtained using a random embedding
instead. Boldface indicates a statistically significant difference ($p<0.05$).}
\end{table}

\begin{table*}[t]
\small
\begin{center}
\resizebox{\textwidth}{!}{
\begin{tabular}{l|ccccccccc|c}
\toprule
& fr             & es    & de    & el    & bg    & ru    & tr    & ar    & vi  & avg.  \\ 
\midrule
\multicolumn{11}{l}{\textit{Selected with English development set}}     \\
\midrule
XLM & 77.3 & 77.9 & 75.9 & 74.3 & 75.3 & 73.8 & 70.4 & 70.9 & 73.2 & 74.3 \\ 
XLM + lang\_emb & \textbf{78.3} & \textbf{79.0} & \textbf{76.5} & \textbf{75.6} & \textbf{76.6} & \textbf{74.8} & \textbf{71.3} & \textbf{72.3} & \textbf{74.4} & \textbf{75.4} \\
\midrule
\multicolumn{11}{l}{\textit{Selected with averaged development set}}     \\
\midrule
XLM & 77.4 & 78.2 & 76.1 & 75.4 & 76.3 & 74.4 & 70.3 & 71.7 & 73.5 & 74.8 \\ 
XLM + lang\_emb & \textbf{78.5} & \textbf{79.0} & \textbf{76.7} & \textbf{75.9} & \textbf{76.8} & \textbf{75.3} & \textbf{71.5} & \textbf{72.4} & \textbf{74.8} & \textbf{75.7}\\
\midrule
XLM-R\textsubscript{Base}  & 77.9 & 78.7 & 76.9 & 76.0 & 77.9 & 75.9 & 72.4 & 72.2 & 74.8 & 75.9 \\
XLM-R\textsubscript{Base} + lang\_emb & \textbf{78.8} & \textbf{79.4} & \textbf{77.4} & \textbf{76.2} & \textbf{78.2} & \textbf{76.1} & \textbf{73.2} & \textbf{72.6} & \textbf{75.4} & \textbf{76.4} \\
\midrule
XLM-R\textsubscript{Large}  & 83.6 & 84.6 & 83.0 & 82.4 & 83.3 & 80.3 & 79.1 & 79.0 & 80.0 & 81.7 \\
XLM-R\textsubscript{Large} + lang\_emb & \textbf{83.9} & \textbf{84.8} & \textbf{83.7} & \textbf{82.8} & \textbf{84.2} & \textbf{81.1} & \textbf{80.3} & \textbf{79.4} & \textbf{80.3} & \textbf{82.3} \\

\bottomrule
\end{tabular}
}
\end{center}
\caption{\label{xnli_results_MUSE} Results on XNLI test set with zero-shot prediction. The results show that adding language embeddings outperforms the baselines in all settings.}
\end{table*}


\subsection{Cross-lingual dependency parsing}

The cross-lingual dependency parsing results in Table~\ref{tab:dependency_results}
indicate that our language embeddings are in fact effective in allowing a parsing
model to leverage information from different languages to parse a new language. 
Substantial accuracy improvements were observed for 13 of the 16 languages used
in the experiment, while accuracy degradation was observed for two languages.
Notably, there were large improvements for each of the four Romance languages used
(ranging from 7.32 to 10.62 absolute points), and a steep drop in accuracy for 
Hebrew (-8.21). Although a sizeable improvement was observed for the only
other language from the same genus in our experiment (Arabic, with a 4.07 improvement), accuracy for the two Semitic languages was far lower than the accuracy for the other genera. This is likely due to the over-representation of Indo-European languages in our dataset, and the lower quality of the MUSE word alignments for these languages (Appendix \ref{muse_word}).

While our accuracy results are well below current results obtained with supervised methods (i.e. using training data for each target language), the average accuracy improvement of 3.4 over the baseline, which uses the exact same parsing setup but without language embeddings, shows that our language embeddings encode actionable syntactic information, corroborating our results using WALS.


\subsection{XNLI prediction}
The XNLI results in Table \ref{xnli_results_MUSE} indicate that our language embeddings, which capture relationships between each test language and the training language (English), are also effective in tasks involving higher-level semantic information. We observe consistent performance gains over very strong baselines in all settings and models for each language. Specifically, in the fully zero-shot setting where we select the best model based on the English development data, adding our learned language embeddings increases 1.1 absolute points on average for XLM.
The same trend holds for XLM-R results, not shown due to space limits. 
On the other hand, if we select the best model on the averaged development set following \citet{conneau-etal-2020-unsupervised}, we observe averaged performance gain of 0.9, 0.5, and 0.6 absolute points for XLM, XLM-R\textsubscript{Base}, and XLM-R\textsubscript{Large}, respectively. We conjecture that the lower improvement on XLM-R models compared to XLM is due to that XLM-R was pretrained without language embeddings. When we add our language embeddings to the original word and positional embeddings, the distribution of the overall input embedding such as variance is changed. Hence, the language embeddings can be considered as noise at the beginning, making it hard to learn and incorporate additional information. However, the improvement is consistent over all strong baselines, suggesting that our language embeddings, which are not optimized towards any specific task, can be leveraged off-the-shelf in large pretrained models and achieve better zero-shot transfer ability in downstream tasks.

\subsection{Discussion}
Our results in each of the intrinsic and extrinsic evaluation settings demonstrate that our denoising autoencoder objective, which has been shown to be effective in various language model pre-training tasks \cite{lewis-etal-2020-bart, raffel-etal-2020-exploring}, is effective for learning language embeddings that capture typological information and can be used to improve cross-lingual inference. Even though reconstructing the original sentence from a randomly ordered string is the direct training objective, our evaluation of the resulting embeddings is not based simply on word order. 

The grammar of a language is of course an important factor in determining the order of words in a sentence in that language, although it is not the only factor. The syntax area features in our WALS evaluation, which are largely related to relative orders of constituents and syntactic constructions and therefore clearly relevant to our training objective, confirm that part of what our embeddings capture is in fact related to word ordering. However, our results on the lexicon and morphology areas indicate that language-specific information capture in our embeddings goes beyond ordering information. Although it may seem that the model only has access to information about word ordering during training, text in the various languages also provides information about word usage, co-occurrence, and to some extent even inflection through the word embeddings. As a result, language embeddings trained with our approach capture interpretable and useful typological information beyond word order. Because language embeddings are the only signal to the model indicating what each of the languages that are mixed within the training data reads like, we conjecture that our denoising autoencoder objective encourages the embeddings to encode language-specific information necessary to distinguish each language from the others.

\section{Conclusion}



Language embeddings have the potential to contribute to our understanding of language and linguistic typology, and to improve the performance of downstream multilingual NLP applications. 
Our proposed method to generate dense vectors to capture language features is relatively simple, based on the idea of denoising autoencoders. The model does not require any labeled or parallel data, which makes it promising for cross-lingual learning in situations where no task-specific training datasets are available. 

We showed that the trained language embeddings represent typological information,
and can also benefit the downstream tasks in a zero-shot learning setting.
This is an encouraging result that indicates that task-specific annotated data for various languages can be leveraged more effectively for improved task performance in situations where language-specific resources may be scarce. At the same time, 
our 
results
indicate that the effectiveness of our approach is sensitive to the set of languages used,
highlighting the importance of using a more balanced variety of languages than is current practice, our work included. We will pursue an 
investigation of the impact of language selection in multilingual and cross-lingual models as future work, to our understanding of these methods and their broader applicability.

\section*{Acknowledgments}
We thank the anonymous reviewers for their constructive suggestions. This work was supported by the National Science Foundation under Grant No. 1840191. Any opinions, findings, and conclusions or recommendations expressed are those of the authors and do not necessarily reflect the views of the NSF.

\section*{Ethical Consideration}
Our motivation to learn language embeddings without parallel data is to understand how language relationships and typology can be generated without any human annotation. 
We also explore how our learned language embeddings can be applied to downstream tasks. We hope that our proposed method can inspire future research on generating and utilizing typology in cross-lingual settings because we may not have a large amount of translation data for each language, which has been widely  used in past research on data-driven  modeling of linguistic typology. Since our proposed method can be easily adapted to different architectures and pre-trained models with minimal cost (in terms of both data annotation cost and computation cost), it can reduce resources needed when applying language embeddings for zero-shot cross-lingual downstream tasks. We run all our experiments on two TITAN RTX GPUs and two RTX 2080Ti GPUs. We compare our language embeddings to baselines in the standard settings in literature.

\bibliographystyle{acl_natbib}
\bibliography{anthology,acl2021}

\newpage
\clearpage

\appendix

\section{Implementation Details} 

\subsection{Denoising autoencoder} \label{training_details}
We use the CommonCrawl dataset from the CoNLL 2017 shared task \cite{conll_2017_auto_data}: 
\url{http://hdl.handle.net/11234/1-1989}.
We implement the denoising autoencoder with a two-layer LSTM with 500 hidden units and global attention \cite{luong-etal-2015-effective} using a modified version of OpenNMT~\citep{klein-etal-2017-opennmt}. 
We use a batch size of 16 and Adam optimization \cite{adam} for training with initial learning rate of 1, 0.85 decay applied every 25,000 steps after the first 10,000 steps. The word embedding size if 300 pretrained from MUSE and the language embedding size is 50. We apply global attention \cite{lu-etal-2018-neural} between the decoder and the encoder.

For experiment with XLM \cite{xlm}, we use the provided code base \footnote{\url{https://github.com/facebookresearch/XLM}} following the suggested preprocessing processes and training details.

\subsection{XNLI} \label{xnli_details}
For XNLI experiments with both XLM and XLM-R, we follow the hyper-parameter tuning suggestions in the code base and author response. We tune the hyper-parameters on the English development set to match the scores reported in the corresponding papers, and use the same hyper-parameters for all runs.

Specifically, for XLM, we fine-tuned the \texttt{mlm\_tlm\_xnli15\_1024} model with the implementation  from the XLM code base \cite{xlm}. We use a learning rate of 5e-6 (from a suggested range of [5e-6, 2.5e-5, 1.25e-4]), a batch size of 8 (from suggested range of [4, 8]), and run 150 epochs (with early stopping if  the validation accuracy does not improve for 5 epochs) where each epoch size is 20000 examples, taking 510s on a single TITAN RTX GPU.

For XLM-R, we modified the Huggingface implementation \cite{wolf-etal-2020-transformers}. We use a learning rate of 7.5e-6, accumulated batch size of 128, and run 10 full epochs (with early stopping).
We evaluate on the development set every 720 training steps. For each epoch, XLM-R base takes 6468s on a single RTX 2080Ti GPU, and XLM-R takes 18306s on a single TITAN RTX GPU.

	

\section{WALS Categories}
\label{app_wals_details}
    \indent \textbf{Lexicon}: 129A Hand and Arm, 138A Tea;
    \indent \textbf{Syntax}: 81A Order of Subject, Object and Verb, 82A Order of Subject and Verb, 83A Order of Object and Verb, 84A Order of Object, Oblique, and Verb, 85A Order of Adposition and Noun Phrase, 86A Order of Genitive and Noun, 87A Order of Adjective and Noun, 88A Order of Demonstrative and Noun, 92A Position of Polar Question Particles, 93A Position of Interrogative Phrases in Content Questions, 95A Relationship between the Order of Object and Verb and the Order of Adposition and Noun Phrase, 96A Relationship between the Order of Object and Verb and the Order of Relative Clause and Noun, 97A Relationship between the Order of Object and Verb and the Order of Adjective and Noun, 106A Reciprocal Constructions, 110A Periphrastic Causative Constructions, 113A Symmetric and Asymmetric Standard Negation, 114A Subtypes of Asymmetric Standard Negation, 121A Comparative Constructions, 122A Relativization on Subjects, 125A Purpose Clauses, 126A 'When' Clauses, 127A Reason Clauses, 128A Utterance Complement Clauses, 144B Position of negative words relative to beginning and end of clause and with respect to adjacency to verb;
    \indent \textbf{Partially Morphological}: 30A Number of Genders, 31A Sex-based and Non-sex-based Gender Systems, 32A Systems of Gender Assignment, 34A Occurrence of Nominal Plurality, 35A Plurality in Independent Personal Pronouns, 36A The Associative Plural, 37A Definite Articles, 38A Indefinite Articles, 41A Distance Contrasts in Demonstratives, 43A Third Person Pronouns and Demonstratives, 44A Gender Distinctions in Independent Personal Pronouns, 45A Politeness Distinctions in Pronouns, 46A Indefinite Pronouns, 47A Intensifiers and Reflexive Pronouns, 48A Person Marking on Adpositions, 49A Number of Cases, 50A Asymmetrical Case-Marking, 51A Position of Case Affixes, 52A Comitatives and Instrumentals, 53A Ordinal Numerals, 54A Distributive Numerals, 57A Position of Pronominal Possessive Affixes, 62A Action Nominal Constructions, 65A Perfective/Imperfective Aspect, 66A The Past Tense, 67A The Future Tense, 68A The Perfect, 71A The Prohibitive, 72A Imperative-Hortative Systems, 74A Situational Possibility, 75A Epistemic Possibility, 76A Overlap between Situational and Epistemic Modal Marking, 77A Semantic Distinctions of Evidentiality, 78A Coding of Evidentiality, 98A Alignment of Case Marking of Full Noun Phrases, 101A Expression of Pronominal Subjects, 102A Verbal Person Marking, 103A Third Person Zero of Verbal Person Marking, 111A Nonperiphrastic Causative Constructions, 112A Negative Morphemes, 115A Negative Indefinite Pronouns and Predicate Negation, 116A Polar Questions, 117A Predicative Possession, 118A Predicative Adjectives, 119A Nominal and Locational Predication, 120A Zero Copula for Predicate Nominals, 124A 'Want' Complement Subjects, 143A Order of Negative Morpheme and Verb, 143F Postverbal Negative Morphemes, 144A Position of Negative Word With Respect to Subject, Object, and Verb, 144D The Position of Negative Morphemes in SVO Languages, 144I SNegVO Order, 144J SVNegO Order, 144K SVONeg Order;
    \indent \textbf{Non-learnable}: 1A Consonant Inventories, 2A Vowel Quality Inventories, 3A Consonant-Vowel Ratio, 4A Voicing in Plosives and Fricatives, 5A Voicing and Gaps in Plosive Systems, 6A Uvular Consonants, 9A The Velar Nasal, 11A Front Rounded Vowels, 12A Syllable Structure, 14A Fixed Stress Locations, 15A Weight-Sensitive Stress, 16A Weight Factors in Weight-Sensitive Stress Systems, 17A Rhythm Types, 19A Presence of Uncommon Consonants, 21A Exponence of Selected Inflectional Formatives, 21B Exponence of Tense-Aspect-Mood Inflection, 22A Inflectional Synthesis of the Verb, 23A Locus of Marking in the Clause, 24A Locus of Marking in Possessive Noun Phrases, 25A Locus of Marking: Whole-language Typology, 26A Prefixing vs. Suffixing in Inflectional Morphology, 27A Reduplication, 29A Syncretism in Verbal Person/Number Marking, 69A Position of Tense-Aspect Affixes, 70A The Morphological Imperative, 79A Suppletion According to Tense and Aspect, 79B Suppletion in Imperatives and Hortatives, 104A Order of Person Markers on the Verb, 136A M-T Pronouns, 142A Para-Linguistic Usages of Clicks.

\section{Additional Results}
\label{app_addtional_results}

\subsection{WALS} 
\label{app_wals}
Additional comparison on WALS with different language embedding baselines.
\begin{table}[h] 
\small
\begin{center}
\resizebox{\columnwidth}{!}{
\begin{tabular}{l|cccc|c}
\toprule
& Lexicon & Syntax & Part. Morph. & Non-learn. & Rand \\
$n$ features & 2 & 13 & 46 & 21 & - \\
\midrule
Random  & 0.56 & 0.61 & 0.52 & 0.52 & - \\
Majority  & 0.68 & 0.76 & 0.68 & 0.68 & - \\
\midrule
XLM mono & 0.68 & 0.76 & 0.64 & 0.67  & 0.11 \\
\midrule
Spe. CC & 0.66 & 0.77\sig & 0.67 & 0.67 & 0.58 \\
Eng. CC & \textbf{0.86\sig} & 0.80\sig & \textbf{0.72\sig} & \textbf{0.70\sig} & \textbf{0.62} \\
Wiki & 0.82\sig & \textbf{0.80\sig} & 0.70\sig & 0.70\sig & \textbf{0.62}\\

\bottomrule
\end{tabular}
}
\end{center}

\caption{\label{app_wals_results} Results on the WALS prediction task and linguistic typology on 29 languages across 10 language genera. 
\sig indicates statistical significance ($p<0.01$) over the Majority baseline.}
\end{table}

\begin{table}[h] 
\small
\begin{center}
\resizebox{\columnwidth}{!}{
\begin{tabular}{l|cccc}
\toprule
& Lexicon & Syntax & Part. Morph. & Non-learn.  \\
$n$ features & 1 & 13 & 38 & 25  \\
\midrule
Eng. & \textbf{0.71} & \textbf{0.61} & 0.53 & \textbf{0.51} \\
XLM parallel & 0.28 & 0.57 & \textbf{0.56} & 0.50\\

\bottomrule
\end{tabular}
}
\end{center}

\caption{\label{app_wals_results_10} Results on the WALS prediction task and linguistic typology on 10 languages in comparison to XLM language embeddings trained from XNLI language parallel data (MLM + TLM objectives).}
\end{table}

\subsection{Cross-lingual dependency parsing}
\begin{table}[h!]
\small
\begin{tabular}{lccc}
\toprule
Language & Spe. & Wiki & XLM mono \\
\midrule
\multicolumn{3}{l}{\textit{Finnic}} \\
\quad Estonian & 60.27 & 61.36 & 53.51\\
\quad Finnish & 62.49 & 62.32 & 55.65 \\
\multicolumn{3}{l}{\textit{Germanic}} \\
\quad Danish & 68.81 & 70.37 & 66.48 \\
\quad English & 73.00 & 69.68 & 70.96 \\
\quad German & 63.64 & 65.49 & 64.01\\
\quad Norwegian & 77.76 & 75.42 & 74.24 \\
\quad Slovene & 71.17 & 71.19 & 64.29 \\
\multicolumn{3}{l}{\textit{Romance}} \\
\quad Catalan & 81.72 & 81.95 & 79.13\\
\quad French & 78.57 & 79.27 & 72.52 \\
\quad Spanish & 83.38 & 82.09 & 81.40\\
\quad Portuguese & 79.65 & 80.09 & 80.40 \\
\multicolumn{3}{l}{\textit{Semitic}} \\
\quad Arabic & 52.58 & 51.26 &  50.58\\
\quad Hebrew & 38.93 & 33.01 & 47.56 \\
\multicolumn{3}{l}{\textit{Slavic}} \\
\quad Bulgarian & 66.22 & 64.57 & 62.09\\
\quad Czech & 67.31 & 59.15 & 64.66\\
\quad Russian & 62.92 & 62.08 & 62.04 \\
\midrule
Average & 68.02 & 66.83 &  65.60\\
\bottomrule
\end{tabular}
\caption{\label{app_dependency_results} Zero-shot parsing results (UAS) comparing Spe., Wiki, and XLM mono language embeddings. Results show that using language embeddings can improve parsing performance, and our methods outperform previous methods by a large margin (2.4 absolute points).}
\end{table}

\onecolumn
\clearpage
\subsection{Linguistic typology} \label{app_topology}
\begin{figure*}[ht!]
\centering
\includegraphics[scale=0.5]{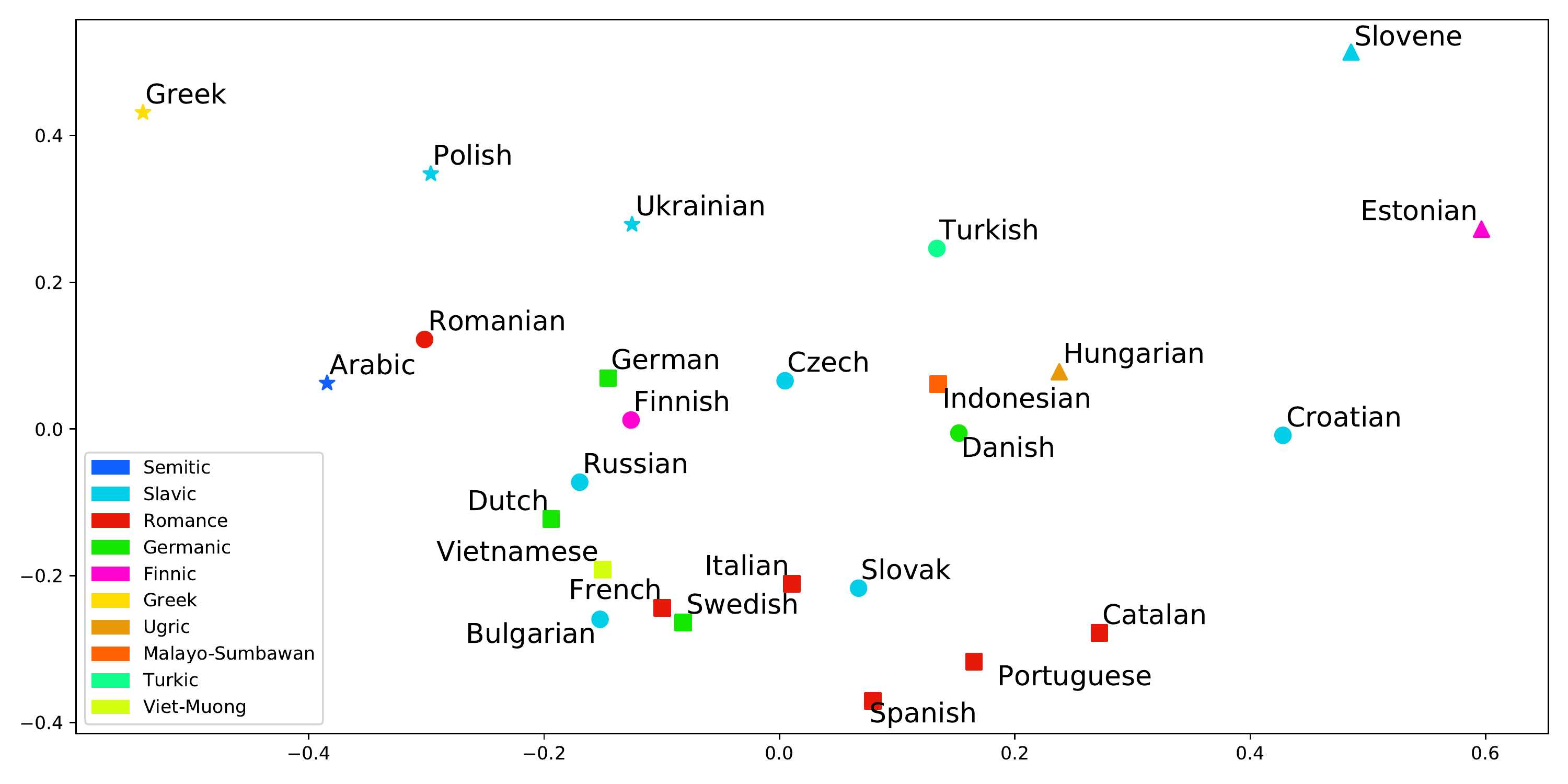}
\caption{\label{pca_bible} Two-dimensional PCA projection of the  language embeddings from \citet{malaviya-etal-2017-learning}. }
\end{figure*}

\subsection{XNLI}
\begin{table*}[h]
\small
\begin{center}
\resizebox{\textwidth}{!}{
\begin{tabular}{l|ccccccccc|c}
\toprule
& fr             & es    & de    & el    & bg    & ru    & tr    & ar    & vi  & avg.  \\ 
\midrule
\multicolumn{11}{l}{\textit{Selected with averaged development set}}     \\
\midrule
XLM-R\textsubscript{Base} + Spe. & 78.9 & 79.8 & 77.2 & 76.0 & 78.2 & 76.2 & 73.8 & 72.5 & 75.3 & 76.4 \\
\midrule
XLM-R\textsubscript{Base} + Wiki & 79.0 & 78.7 & 76.4 & 76.1 & 78.3 & 75.9 & 73.8 & 71.8 & 75.5 & 76.2 \\
\midrule
XLM-R\textsubscript{Base} + XLM mono & 78.2 & 78.4 & 76.7 & 76.4 & 78.1 & 75.9 & 73.6 & 72.3 & 75.1 & 76.1 \\
\midrule
XLM-R\textsubscript{Base} + XLM parallel & 78.3 & 78.9 & 76.5 & 76.6 & 77.6 & 75.4 & 73.2 & 72.4 & 75.0 & 76.0 \\

\bottomrule
\end{tabular}
}
\end{center}
\caption{\label{app_xnli_results} Results on XNLI test set with zero-shot prediction comparing different language embeddings on XLM-R\textsubscript{Base}. We cannot compare directly to machine translation-based methods such as \citet{malaviya-etal-2017-learning} because there is no English embeddings. The results show that on XLM-R\textsubscript{Base}, our language embeddings perform better than language embeddings from previous research.}
\end{table*}

\section{MUSE Word Translation Accuracy}
\label{muse_word}
\begin{table*}[h]
\small
\begin{center}
\resizebox{\textwidth}{!}{
\begin{tabular}{l|ccccccccccc}
\toprule
& en     & fr    & es    & de    & el    & bg    & ru    & tr   & he   & ar    & vi   \\
\midrule
en & 100.00 & 83.00 & 83.85 & 77.07 & 59.81 & 60.86 & 67.51 & 61.48 & 58.69  & 52.04 & 54.57 \\
\bottomrule
\end{tabular}
}
\caption{Precision at k = 1 word translation to English for the most frequent 50,000 words in each language using CSLS for the generated dictionary.} \label{word_translation_accuracy}
\end{center}
\end{table*}

\end{document}